\documentclass[runningheads]{llncs}
\usepackage[T1]{fontenc}
\usepackage{graphicx,verbatim}
\begin{document}
\title{Ask4VG: Risk-Aware Question Selection for Reducing Prior-Driven Answers in Medical VQA}

\author{Xiaorong Zhu\inst{1} \and Qiang Li\inst{1} \and Zibo Xu\inst{1} \and Weijie Wang \inst{2} \and Weizhi Nie\thanks{Corresponding author.}\inst{3}}

\authorrunning{X. Zhu et al.}

\institute{School of Microelectronics, Tianjin University, Tianjin 300072, China 
 \and
DISI, University of Trento, Trento, Italy
 \and
School of Electrical and Information Engineering, Tianjin University, Tianjin 300072, China\\
\email{weizhinie@tju.edu.cn}}
  
\maketitle              
\begin{abstract}
Medical visual question answering requires models to ground their responses in image evidence, because visually unsupported answers can mislead downstream interpretation. However, many medical VQA questions are generic, template-like, or highly similar in form, which can encourage models to learn question-answer shortcuts instead of image-dependent reasoning and thereby increase the risk of hallucinated responses. We propose Ask4VG, a label-free pilot framework for risk-aware question selection. Ask4VG estimates question-induced hallucination risk through counterfactual visual probing: the same question is asked under the original image, a perturbed image, a blank image, and a mismatched image, and the resulting answer relations are converted into weak supervision for a counterfactual risk estimator. The learned estimator then reranks candidate question rewrites to favor intent-preserving questions that are less invariant to missing or mismatched visual evidence before final answer generation. On VQA-RAD with Qwen2-VL-2B-Instruct, prompt-only rewriting increases counterfactual risk, whereas predicted-risk reranking reduces held-out risk from 0.658 to 0.623 and improves exact accuracy from 0.337 to 0.356. A 300-sample PMC-VQA external check shows the same direction of risk reduction with a small accuracy gain. These results suggest that question selection is a promising complement to response-level hallucination mitigation for reliable medical VQA.


\end{abstract}
\section{Introduction}

Medical visual question answering (Medical VQA) aims to answer clinical questions from medical images and is increasingly viewed as an important interface for multimodal medical reasoning~\cite{tmi}. Unlike general-domain VQA, medical VQA requires stronger visual grounding: an answer should be supported by image evidence rather than by plausible language patterns alone. However, reliable medical VQA remains challenging. Many questions in existing datasets are generic, repetitive, or similar in form, such as questions about whether an abnormality is present, which organ system is affected, or whether a finding is normal. Such question patterns may encourage vision-language models to learn question-answer shortcuts during training or inference~\cite{lmd1,lmd2,zc}. As a result, a model may produce plausible but visually unsupported answers, increasing hallucination risk in clinically sensitive scenarios.

While recent advances in medical datasets \cite{lau2018vqarad,liu2021slake,he2020pathvqa,zhang2023pmcvqa} and biomedical VLMs \cite{li2023llavamed,moor2023medflamingo} show promise, most hallucination mitigation strategies focus on post-hoc answer correction. In this work, we explore a complementary pre-generation perspective: treating the question formulation as a controllable factor. Because different phrasings of the same clinical intent induce varying levels of image dependence, a vague question may trigger language priors, whereas a carefully constructed rewrite can force closer visual attention. By estimating this question-induced hallucination risk, we can guide the selection of intent-preserving questions that minimize prior-driven shortcuts before generation.

Based on this motivation, we propose Ask4VG, a risk-aware question selection framework for reducing prior-driven answers in Medical VQA. The core of Ask4VG is a label-free hallucination risk estimation function based on counterfactual visual probing. Given an image-question pair, we query a frozen vision-language model under multiple visual conditions, including the original image, a perturbed image, a blank image, and a mismatched image. If the model produces similar answers even when the visual evidence is missing or replaced, the question is considered more likely to induce prior-driven reasoning. These counterfactual answer relations are converted into weak supervision for a hallucination risk estimator. The learned estimator is then used as feedback to guide question selection: candidate rewrites are ranked according to predicted risk while preserving the original clinical intent and answerability. In this way, Ask4VG shifts part of hallucination mitigation from response-level correction to pre-generation question optimization.

Our contributions are three-fold. 
\begin{itemize}
    \item We formulate question-induced hallucination risk as a measurable factor in Medical VQA and introduce a counterfactual risk estimation function that captures visual insensitivity, question-only bias, and mismatch invariance.
    \item We propose a risk-feedback question selection strategy that uses the estimated risk to choose intent-preserving, visually grounded question rewrites before final answer generation.
    \item We conduct experiments on VQA-RAD and a small external PMC-VQA check, showing that risk-guided question selection reduces counterfactual hallucination risk and yields modest but consistent accuracy gains compared with original questions and prompt-only rewriting.
\end{itemize} 

\section{Related Work}
Medical VQA requires models to answer clinically relevant questions from medical images, making visual grounding especially important\cite{lau2018vqarad,liu2021slake,he2020pathvqa,zhang2023pmcvqa,li2023llavamed,moor2023medflamingo}
. VQA-RAD introduced clinician-generated questions and answers for radiology images~\cite{lau2018vqarad}, while PMC-VQA scaled medical VQA through visual instruction tuning from biomedical figures~\cite{zhang2023pmcvqa}. These datasets provide useful testbeds for medical multimodal reasoning, but their questions often contain repeated templates or similar linguistic forms, which may encourage models to exploit question-answer regularities rather than image evidence.

Language priors and question biases have long been recognized in VQA\cite{antol2015vqa,goyal2017making,agrawal2018dont,hudson2019gqa,gurari2018vizwiz}
. VQA v2 was proposed to reduce language bias by pairing similar questions with complementary images and answers~\cite{goyal2017making}, and VQA-CP further exposed the brittleness of models that rely on answer priors rather than visual evidence~\cite{agrawal2018dont}. Our work follows this line of motivation but studies it in medical VQA, where question-induced shortcuts may increase hallucination risk.

Recent studies show that large vision-language models can generate visually unsupported answers\cite{rohrbach2018object,dai2023plausible,li2023evaluating,fu2023mme,guan2024hallusionbench}
. POPE evaluates object hallucination through polling-based questions~\cite{li2023evaluating}, and visual contrastive decoding mitigates hallucination by contrasting outputs under different visual conditions~\cite{leng2024mitigating}. These methods mainly evaluate or intervene at the answer-generation stage. In contrast, Ask4VG treats the question as a controllable factor and uses counterfactual visual probing to estimate hallucination risk before final answer generation.

\section{Method}

\begin{figure*}[t]
    \centering
    \includegraphics[width=\textwidth]{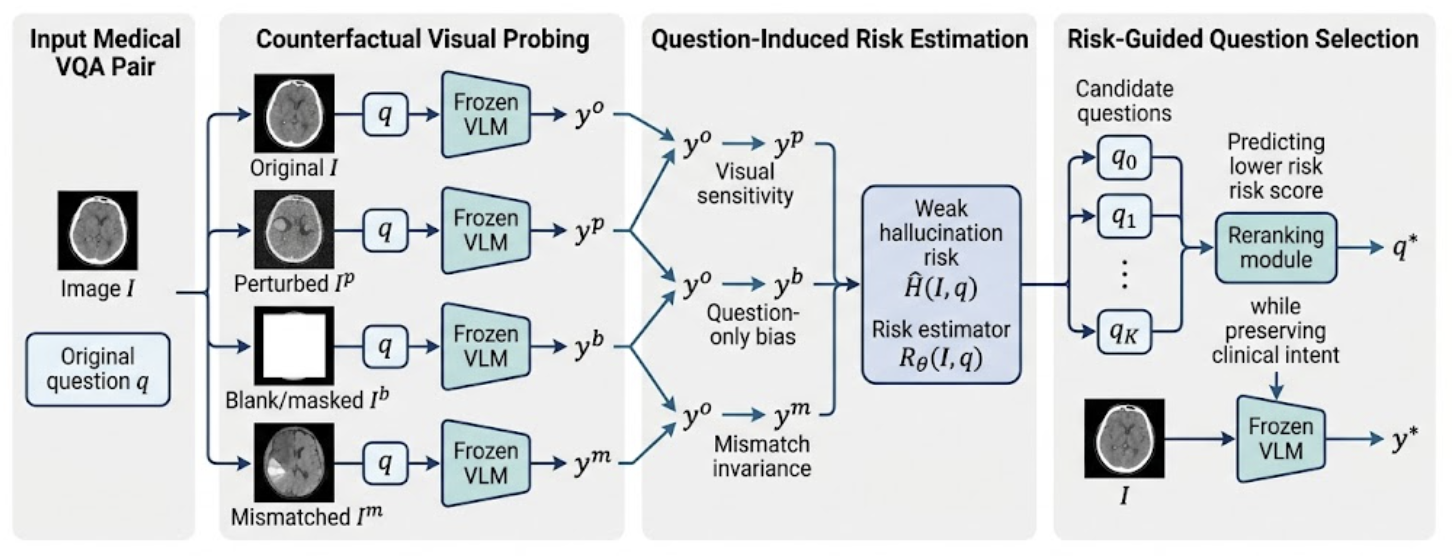}
    \caption{
Overview of the Ask4VG framework. Counterfactual visual probing assesses question-induced hallucination risk across four image conditions (original, perturbed, blank, mismatched). The extracted risk signals then guide the selection of an intent-preserving, low-risk question prior to final answer generation.
    }
    \label{fig:framework}
\end{figure*}

\subsection{Problem Formulation}
Given a medical image $I$ and an initial question $q$, a vision-language model $f_{\psi}$ generates an answer
\begin{equation}
    y = f_{\psi}(I, q).
\end{equation}
In medical VQA, the desired answer should be grounded in visual evidence from $I$. However, some questions may be generic, template-like, or weakly image-dependent, allowing the model to produce plausible answers from question-answer priors. We therefore define a question-induced hallucination risk function
\begin{equation}
    R_{\theta}(I, q) \in [0,1],
\end{equation}
where a higher value indicates that answering $q$ is more likely to rely on language priors or dataset shortcuts rather than image evidence. Our goal is to select a revised question $q^{*}$ from a candidate set $\mathcal{C}$ such that it preserves the clinical intent of $q$ while reducing the estimated hallucination risk:
\begin{equation}
    q^{*} = \arg\min_{q' \in \mathcal{C}} R_{\theta}(I, q').
\end{equation}

\subsection{Counterfactual Visual Probing}

To bypass the need for expensive expert hallucination annotations, we estimate risk via label-free counterfactual visual probing~\cite{ldc}. For each image-question pair $(I,q)$, we query the model under four conditions: the original ($I$), perturbed ($I^{p}$), blank ($I^{b}$), and mismatched ($I^{m}$) images, yielding answers $y^{k}=f_{\psi}(I^{k},q)$ for $k \in \{o, p, b, m\}$.

Visually grounded questions should yield image-dependent answers; invariance to missing or altered visuals indicates prior-driven hallucination. We quantify this using semantic distance $D$ and similarity $\mathrm{Sim}$ to compute visual sensitivity ($S_{\mathrm{vis}}$), question-only bias ($S_{\mathrm{prior}}$), and mismatch invariance ($S_{\mathrm{mis}}$):
$$ S_{\mathrm{vis}} = D(y^{o}, y^{p}), \quad S_{\mathrm{prior}} = \mathrm{Sim}(y^{o}, y^{b}), \quad S_{\mathrm{mis}} = \mathrm{Sim}(y^{o}, y^{m}). $$

If a ground-truth answer $a$ is available, we also compute answer consistency $S_{\mathrm{cons}} = \mathrm{Cons}(y^{o}, a)$. The final weak hallucination risk proxy is a weighted combination of these counterfactual signals:
$$ \hat{H}(I,q) = \alpha(1-S_{\mathrm{vis}}) + \beta S_{\mathrm{prior}} + \gamma S_{\mathrm{mis}} + \delta(1-S_{\mathrm{cons}}), $$
where $\alpha, \beta, \gamma, \delta \geq 0$.

\subsection{Risk Estimation}

We train a risk estimator $R_{\theta}$ to predict $\hat{H}(I,q)$ from the image, question, and counterfactual answer relations:
\begin{equation}
    R_{\theta}(I,q) =
    g_{\theta}(e_I, e_q, \Delta_{op}, \Delta_{ob}, \Delta_{om}),
\end{equation}
where $e_I$ and $e_q$ are image and question embeddings, and $\Delta_{op}$, $\Delta_{ob}$, and $\Delta_{om}$ encode answer differences between the original answer and the perturbed, blank, and mismatched answers. The estimator is optimized using a regression loss:
\begin{equation}
    \mathcal{L}_{\mathrm{risk}}
    =
    \left(R_{\theta}(I,q)-\hat{H}(I,q)\right)^2.
\end{equation}
When multiple question candidates are available for the same image, we also use pairwise ranking supervision. If $\hat{H}(I,q_i)<\hat{H}(I,q_j)$, the model is encouraged to assign lower predicted risk to $q_i$ than to $q_j$:
\begin{equation}
    \mathcal{L}_{\mathrm{rank}}
    =
    \max(0, m + R_{\theta}(I,q_i)-R_{\theta}(I,q_j)),
\end{equation}
where $m$ is a margin. The final objective is
\begin{equation}
    \mathcal{L}
    =
    \mathcal{L}_{\mathrm{risk}}
    + \lambda \mathcal{L}_{\mathrm{rank}}.
\end{equation}

\subsection{Risk-Guided Question Selection}

At inference time, Ask4VG generates a small candidate set $\mathcal{C}=\{q_0,q_1,\ldots,q_K\}$, where $q_0$ is the original question and the remaining candidates are produced by a lightweight question rewriting prompt. The prompt asks the model to preserve the original clinical intent while making the question more visually specific.

For each candidate $q' \in \mathcal{C}$, Ask4VG predicts its hallucination risk using $R_{\theta}(I,q')$. To avoid selecting questions that are low-risk but clinically drifted or unanswerable, we use a utility-aware score:
\begin{equation}
    \mathrm{Score}(I,q,q')
    =
    -R_{\theta}(I,q')
    + \lambda_u U(q')
    + \lambda_s \mathrm{Sim}(q,q')
    - \lambda_d D_{\mathrm{deg}}(q'),
\end{equation}
where $U(q')$ measures answerability and specificity, $\mathrm{Sim}(q,q')$ preserves the original intent, and $D_{\mathrm{deg}}(q')$ penalizes degenerate rewrites such as vague, overly long, or repetitive questions. The selected question is
\begin{equation}
    q^{*}
    =
    \arg\max_{q' \in \mathcal{C}}
    \mathrm{Score}(I,q,q').
\end{equation}
The final answer is then generated by the frozen VLM using the selected question:
\begin{equation}
    y^{*}=f_{\psi}(I,q^{*}).
\end{equation}

The framework does not modify the medical image or fine-tune the base VLM. Instead, it uses counterfactual risk feedback to choose a safer question formulation before answer generation.

\section{Experiments and Results}

\subsection{Experimental Setup}
We evaluate on 1,793 image-question-answer pairs from VQA-RAD~\cite{lau2018vqarad} (split 70/15/15\% for training/validation/test by sample ID), with a 300-sample external check on PMC-VQA~\cite{zhang2023pmcvqa}. All experiments use a frozen Qwen2-VL-2B-Instruct~\cite{radford2021clip,alayrac2022flamingo,liu2023llava,li2023blip2,bai2023qwenvl,wang2024qwen2vl,chen2024internvl} with greedy decoding (maximum 16 new tokens). Ask4VG uses counterfactual probing purely to select the optimal question for inference, requiring no model fine-tuning.
We compare several question policies: \textit{Original} dataset questions; \textit{Prompt rewrite} (unconditional rewriting without risk feedback); \textit{Risk-only rerank} (selection guided by Ask4VG); and constrained variants (\textit{Risk+margin}, \textit{Risk+intent/length}, \textit{Utility score}) designed to prevent semantic drift. Finally, we report an undeployable \textit{Oracle bound} based on measured test-time risk as a diagnostic upper limit.

\subsection{Counterfactual Probing Reveals Prior-Driven Question Risk}

We first examine whether medical VQA questions exhibit prior-driven behavior under counterfactual visual conditions. For each image-question pair, we query the frozen VLM using the original image, a perturbed image, a blank or masked image, and a mismatched image sampled from another case. We then compute answer similarities between the original answer and the counterfactual answers. High original-blank similarity suggests question-only bias, whereas high original-mismatch similarity suggests weak image-instance dependence.

Table~\ref{tab:diagnostic} shows the full-dataset diagnostic analysis on VQA-RAD. Original questions already exhibit high original-perturbed and original-mismatch similarity, indicating that many answers remain stable even when visual evidence is degraded or replaced. Prompt-only rewriting changes most questions, but it increases original-blank similarity from 0.4541 to 0.5391 and increases pseudo risk from 0.6558 to 0.6972. This suggests that rewriting questions without risk feedback may amplify, rather than reduce, prior-driven answering. In contrast, measured-risk selection reduces blank and mismatch similarity and lowers pseudo risk, showing that harmful rewrites can in principle be rejected.

\begin{table}[t]
\centering
\caption{Full-dataset diagnostic analysis on VQA-RAD. Measured-risk selection uses measured counterfactual risk and is included only as a diagnostic upper bound. Lower similarity and risk values indicate weaker prior-driven behavior.}
\label{tab:diagnostic}
\begin{tabular}{lccccc}
\\
\hline
Question Strategy & Acc. & Orig-Pert & Orig-Blank & Orig-Mismatch & Risk \\
\hline
Original Question & 0.3475 & 0.8498 & 0.4541 & 0.6974 & 0.6558 \\
Prompt Rewrite & 0.3168 & 0.8625 & 0.5391 & 0.7245 & 0.6972 \\
Measured-Risk Selection & 0.3592 & 0.7943 & 0.3638 & 0.6459 & 0.5976 \\
\hline
\end{tabular}
\end{table}

\subsection{Learning to Predict Question-Induced Risk}

We next evaluate whether weak counterfactual risk can be predicted from observed image-question-answer relations. Table~\ref{tab:risk_prediction} reports held-out VQA-RAD test results for several risk estimators. The question-only model obtains low correlation with measured risk, suggesting that lexical question features alone are insufficient for estimating hallucination risk. Models using only the original answer or counterfactual metadata remain limited when direct answer similarity features are excluded. The full counterfactual risk estimator achieves much stronger correlation with the weak risk target, with a Pearson correlation of 0.816, a Spearman correlation of 0.913, an AUC of 0.999, and a pairwise ranking accuracy of 0.824.

These results should be interpreted as calibration of a weak counterfactual grounding proxy, not as expert-level hallucination recognition. Nevertheless, they support the central assumption of Ask4VG: counterfactual answer relations provide useful feedback for estimating whether a question is likely to induce prior-driven answering.

\begin{table}[t]
\centering
\caption{Risk prediction on the held-out VQA-RAD test split. The full counterfactual model predicts the weak risk proxy more accurately than question-only or reduced-feature baselines.}
\label{tab:risk_prediction}
\resizebox{0.5\textwidth}{!}{\begin{tabular}{lcccc}
\\
\hline
Model & Pearson & Spearman & AUC & Pair Acc. \\
\hline
Question-only & 0.308 & 0.276 & 0.524 & -- \\
Original answer & 0.395 & 0.406 & 0.584 & -- \\
CF meta., no sim. & 0.390 & 0.425 & 0.600 & -- \\
Full CF Risk & 0.816 & 0.913 & 0.999 & 0.824 \\
\hline
\end{tabular}}
\end{table}

\subsection{Risk-Guided Question Selection on VQA-RAD}

We then evaluate whether predicted risk can guide question selection without using measured test-time risk. Table~\ref{tab:vqarad_main} reports non-oracle results on the held-out VQA-RAD test split. Prompt-only rewriting slightly improves exact accuracy but increases pseudo risk from 0.658 to 0.696, indicating that visually phrased rewrites are not necessarily safer. In contrast, risk-only reranking reduces risk to 0.623 and improves exact accuracy from 0.337 to 0.356. The rewrite rate is 20.7\%, suggesting that Ask4VG acts as a selective intervention rather than rewriting every question.
Utility-aware variants further reduce the rewrite rate by requiring sufficient risk reduction or by preserving intent and length constraints. These variants obtain similar accuracy with slightly higher risk than risk-only reranking, reflecting a trade-off between aggressive risk reduction and conservative question preservation. Overall, the results suggest that predicted risk can serve as useful pre-generation feedback for selecting safer question formulations.

\begin{table}[t]
\centering
\caption{Non-oracle question selection on the held-out VQA-RAD test split. Risk-guided selection reduces pseudo risk and yields modest accuracy gains over the original questions and prompt-only rewriting.}
\label{tab:vqarad_main}
\resizebox{0.4\textwidth}{!}{\begin{tabular}{lccc}
\\
\hline
Strategy & Risk & Acc. & Rewrite \\
\hline
Original & 0.658 & 0.337 & 0.0\% \\
Prompt Rewrite & 0.696 & 0.341 & 100.0\% \\
Risk-only Rerank & 0.623 & 0.356 & 20.7\% \\
Risk + Margin & 0.624 & 0.352 & 15.2\% \\
Risk + Intent/Length & 0.633 & 0.356 & 16.3\% \\
Utility Score & 0.625 & 0.352 & 14.4\% \\
Oracle Bound & 0.606 & 0.352 & -- \\
\hline
\end{tabular}}
\end{table}

\subsection{External Check on PMC-VQA}

To reduce the concern that the observed effect is specific to VQA-RAD, we conduct a small external check on 300 held-out PMC-VQA samples. The risk estimator trained on VQA-RAD is applied directly to PMC-VQA question candidates. We treat this experiment as an external sanity check rather than a full cross-dataset benchmark, because the dataset format and answer space differ from VQA-RAD.
As shown in Table~\ref{tab:pmc}, the same trend is observed. Prompt-based visual rewriting increases risk from 0.6483 to 0.6590, whereas risk-guided reranking reduces risk to 0.6200 and improves accuracy from 0.3767 to 0.3933. The utility score obtains the same risk and accuracy with a much lower rewrite rate, suggesting that conservative risk-aware selection can preserve most original questions while still removing a subset of higher-risk ones.

\begin{table}[t]
\centering
\caption{External check on 300 held-out PMC-VQA samples using the VQA-RAD-trained risk estimator. This result is used as a sanity check rather than a full cross-dataset benchmark.}
\label{tab:pmc}
\begin{tabular}{lccc}
\\
\hline
Strategy & Risk & Acc. & Rewrite \\
\hline
Original & 0.6483 & 0.3767 & 0.0\% \\
Visual Rewrite & 0.6590 & 0.3867 & 100.0\% \\
Risk Rerank & 0.6200 & 0.3933 & 39.0\% \\
Utility Score & 0.6200 & 0.3933 & 6.7\% \\
Oracle Bound & 0.6040 & 0.3867 & -- \\
\hline
\end{tabular}
\end{table}

\subsection{Qualitative Analysis}

Table~\ref{tab:qualitative} shows representative examples of successful and harmful rewrites. Successful rewrites typically make the visual evidence explicit, for example by adding ``visible in the image'' or specifying the imaging modality. These changes reduce counterfactual invariance while preserving the original clinical intent. However, some rewrites are superficial or can amplify prior-driven behavior, especially for generic yes/no questions. This supports the need for risk feedback: question rewriting alone is insufficient, and a selection mechanism is needed to reject harmful candidates.

\begin{table}[t]
\centering
\caption{Representative successful and harmful rewrites. Lower risk indicates weaker counterfactual invariance and less prior-driven behavior.}
\label{tab:qualitative}
\resizebox{\linewidth}{!}{
\begin{tabular}{lllc}
\\
\hline
Case & Original Question & Rewritten Question & Risk Change \\
\hline
Success & how large is the lesion? &
What is the size of the lesion visible in the image? & $\downarrow$ \\
Success & is the mass hyperdense? &
Is the mass hyperdense on the CT image? & $\downarrow$ \\
Success & what is the primary abnormality in this image? &
What is the main abnormality visible in this image? & $\downarrow$ \\
Harmful & are the lungs normal appearing? &
Are the lungs on this chest X-ray normal? & $\uparrow$ \\
Harmful & which organ system is abnormal in this image? &
Which organ system is abnormal in this image? & $\uparrow$ \\
\hline
\end{tabular}}
\end{table}

\section{Conclusion}
We presented Ask4VG, a label-free pilot framework for risk-aware question selection in medical VQA. Instead of treating hallucination only as a response-level problem, Ask4VG considers the question itself as a controllable factor that may induce prior-driven answers. By using counterfactual visual probing under original, perturbed, blank, and mismatched images, Ask4VG estimates question-induced hallucination risk and uses this feedback to select more visually grounded, intent-preserving questions before final answer generation. Experiments on VQA-RAD and a small PMC-VQA external check show that prompt-only rewriting can increase counterfactual risk, whereas risk-guided selection reduces risk and yields modest accuracy gains. These findings suggest that question selection is a promising complementary direction for improving the reliability of medical VQA systems. Future work will incorporate expert-rated clinical groundedness, broader medical VLMs, and more diverse medical imaging datasets.

%
%
%
%
\bibliographystyle{splncs04}
\bibliography{refs}

\end{document}